%% file: main.tex
\documentclass{jingdong}
\input{preamble}

\title{Collect for Coverage, Replay with Evidence: Adaptive and Efficient On-Policy Distillation for Long-Horizon Agents}
\author[1,2,3]{Haowen Hou}
\author[3]{Gen Li}
\author[2]{Ruilin Li}
\author[2]{Yicheng Liu}
\author[2]{Zhen Huang}
\author[2]{Shuai Dong}
\author[3]{Congcong Wang}
\author[3]{Kaiwen Tuo}
\author[3]{Hongrui Wu}
\author[3]{Zelin Peng}
\author[3]{Kanghui Tian}
\author[2]{Ji Zhao}
\author[2]{Dianyi Wang}
\author[2,3,\dagger]{Jiaqi Wang}

\affiliation[1]{Shanghai Jiao Tong University}
\affiliation[2]{Shanghai Innovation Institute}
\affiliation[3]{JD.com}
\contribution[\dagger]{Corresponding author}

\abstract{
Fully online on-policy distillation (OPD) is costly for long-horizon agents because each update depends on fresh environment rollouts. Prefix replay offers an efficient alternative by sampling current-student responses from stored histories, reducing idle time for long-tail rollouts and allowing each collected trajectory to support multiple updates. However, our analysis of the complete prefix replay pipeline identifies two limitations in long-horizon settings. In trajectory collection, teacher-only trajectories miss student-induced histories, while broader history coverage can reduce teacher reliability. In trajectory reuse, fixed position-based schedules cannot identify critical decisions that may occur at any trajectory depth. We introduce CARE-OPD (Collect for coverAge, Replay with Evidence) to address both limitations. Student-seeded teacher completions and response-level confidence gating expand history coverage while filtering unreliable updates. Hindsight replay allocation then uses downstream execution evidence to select consequential turns. Across long-horizon tasks Software Engineering and Deep Research, CARE-OPD outperforms online OPD and prior prefix replay baselines. On SWE-bench Verified, it raises task success to 59.5\% while reducing end-to-end training time by 45\% and environment actions by 44\% relative to asynchronous online OPD.
}

\begin{document}
\maketitle

\section{Introduction}
\label{sec:introduction}

On-policy distillation (OPD) trains a student on its own responses while a stronger teacher supplies dense token-level targets~\citep{gu2024minillm,agarwal2024onpolicy}. Fully online OPD for interactive agents executes the student in the environment~\citep{yao2023react}, so training covers histories induced by the student's actions. This alignment becomes expensive over long horizons spanning tens or hundreds of agent turns. A rollout may search a repository, issue web queries, and wait for external tools, so trajectory durations can differ substantially. Synchronous training batches wait for their slowest trajectories, whereas asynchronous pipelines reduce synchronization idle time but still retain the full interaction cost and can introduce stale-policy data~\citep{kang2026asyncopd,chen2026fopd}. Long and heterogeneous rollouts therefore make training much slower than the model update alone.

Prefix replay offers an off-environment alternative. A separate collection stage records multi-turn trajectories from fixed policy snapshots. During training, the learner selects a stored prefix, generates a response of only one turn from the current student, and evaluates the teacher along that response. The generated response is not executed in the environment. Prefix replay therefore removes variable-length environment rollouts from the learner's critical path and allows one executed trajectory to support repeated current-student updates.

Prefix replay reduces training cost by constructing updates from stored histories rather than fresh environment rollouts. The quality of these updates depends on two design choices: \textbf{which histories to collect so that replay reflects situations the student may face while retaining reliable teacher supervision, and which turns to replay so that training focuses on decisions worth learning}. Concurrent work ReOPD collects teacher-generated histories and uses a monotone position-decay schedule to favor early prefixes~\citep{liao2026reopd}. This design makes both parts of replay static: collection fixes the history source to the teacher, while allocation within each collected trajectory depends only on turn position. These choices are restrictive for long-horizon agents, where early actions can alter many later turns, and consequential decisions can occur at any depth. We address the resulting collection and reuse challenges in turn.

The first challenge is to pair student-relevant histories with confident teacher supervision. Teacher-only trajectories omit states reached after student errors, and this history mismatch compounds over long horizons~\citep{ross2011reduction,li2026dagger}. To cover situations that the student may face, a frozen student generates an initial segment before the teacher completes the trajectory. These student-induced histories make the replay contexts more relevant to the student, but they may fall outside the teacher's usual interaction distribution. We therefore pair student-seeded teacher completions with response-level confidence gating: collection broadens coverage toward student-induced histories, while the gate retains responses for which the teacher provides a confident target.

The second challenge is to identify which turns in a long trajectory are worth replaying. Long trajectories interleave routine interactions with consequential diagnoses, corrections, and verification decisions~\citep{li2026criticalstep}. Replaying every turn spends training computation on steps with limited learning value, while position decay can suppress valuable decisions that occur late in the trajectory~\citep{liao2026reopd}. We instead allocate replay after observing the completed trajectory, when later execution feedback and the final outcome can support the value of earlier decisions. After each rollout, we append a selection query to the rollout teacher and reuse its trajectory KV cache to select these evidence-supported turns efficiently.

Together, these components form CARE-OPD (Collect for coverAge, Replay with Evidence). CARE-OPD addresses trajectory collection and reuse through reliability-controlled history coverage and hindsight replay allocation.
Under matched hardware, CARE-OPD reduces end-to-end training time by 45\% and executes 56\% as many environment interactions as fully online OPD. On two long-horizon tasks SWE and Deep Research, it improves task success over all prefix replay and fully online OPD methods. We make three contributions:

\textbf{(1)} We analyze the complete prefix replay pipeline as two design problems: collecting student-relevant histories with reliable teacher supervision, and identifying which decisions in a long trajectory are worth replaying.

\textbf{(2)} We develop an adaptive replay method that addresses both problems. Student-seeded teacher completions cover histories induced by student behavior, while response-level confidence gating filters uncertain teacher supervision. Downstream execution evidence then directs replay toward consequential diagnoses, corrections, and verification decisions.

\textbf{(3)} We evaluate the complete training process on Software Engineering and Deep Research tasks. The method improves task performance over fully online OPD and prefix replay methods while reducing wall-clock time and environment interactions. Therefore, we establish prefix replay as a practical environment-decoupled training regime for long-horizon agents.

\section{Related Work}
\label{sec:related-work}

\subsection{Efficient Multi-Turn On-Policy Distillation}

On-policy distillation (OPD) trains a student on its own responses using a teacher's next-token distribution~\citep{gu2024minillm,agarwal2024onpolicy,dong2026rl}. Recent extensions adapt OPD to multi-turn agents through teacher intervention, budget-aware rollout and loss allocation, hybrid objectives, and asynchronous rollout pipelines~\citep{li2026guidedopd,zhou2026turnopd,tan2026atod,kang2026asyncopd,chen2026fopd}. These methods retain live environment execution during student training.

Prefix replay collects histories independently and trains on current-student responses generated from stored prefixes, keeping environment execution off the training process. Concurrent work ReOPD reuses prefixes from teacher-generated trajectories and assigns each turn a sampling probability that decreases monotonically with its position in the trajectory, thereby favoring earlier turns~\citep{liao2026reopd}. We find that this step-decay schedule cannot effectively transfer to long-horizon tasks spanning tens or hundreds of turns.
We analyze the complete prefix replay pipeline for long-horizon tasks and identify the limitations in both stages, trajectory collection and reuse. We deal with both: student-seeded teacher completions expand history coverage under response-level confidence control, while downstream execution evidence allocates replay across the completed trajectory.

\subsection{Selecting Histories and Updates in OPD}

OPD depends on both the teacher target and the histories where the teacher is queried. Fully online OPD uses student-induced histories, while Guided-OPD mixes teacher- and student-generated turns through an intervention curriculum~\citep{li2026guidedopd}. TrOPD restricts OPD to regions with reliable teacher supervision and uses teacher prefixes for off-policy guidance~\citep{xing2026tropd}. These methods control the tension between student-relevant histories and teacher reliability through online intervention and trust regions. CARE-OPD instead constructs a replay pool with student-seeded teacher completions, then reassesses teacher confidence after the current student samples a response.

Several OPD methods also select or weight supervision within a trajectory. TurnOPD allocates rollout and loss budgets using turn-level statistics, SAGE-OPD selects teacher interventions from environment feedback and weights supervision by teacher confidence, and ATOD reweights turns using teacher--student disagreement and uncertainty~\citep{zhou2026turnopd,zhou2026sageopd,tan2026atod}. Other work masks or reweights OPD supervision using self-uncertainty, token teachability, or position-dependent teacher reliability~\citep{ke2026egrsd,wang2026teachability,liu2026teacherreliability}. These rules derive their decisions from online trajectories, model distributions, or token position. Our replay setting adds an allocation decision over stored prefixes: evidence from the completed trajectory determines where the replay loss is evaluated, and teacher confidence on the newly sampled response determines whether that loss contributes an update.

\begin{figure}[!htbp]
    \centering
    \includegraphics[width=\textwidth]{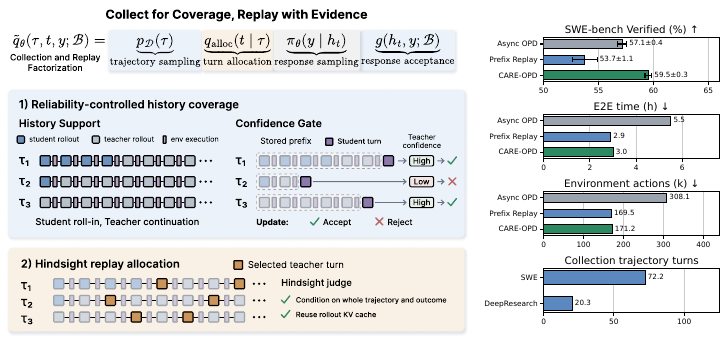}
    \caption{Overview of CARE-OPD. \emph{Left:} Factorization of prefix replay in agent OPD (above). Method overview (below). \emph{Right:} The plots compare SWE-bench Verified task success, end-to-end time, and environment actions on Qwen3.5 family. The bottom panel reports mean collection trajectory length on SWE and Deep Research.  Async OPD denotes online asynchronous one-turn distillation, and Prefix Replay denotes vanilla prefix replay.}
    \label{fig:method-overview}
\end{figure}

\section{Method}
\label{sec:method}

In this section, we first define agentic OPD and prefix replay, then analyze the whole pipeline and identify two limitations of static prefix policies. CARE-OPD addresses them through reliability-controlled history coverage and hindsight replay allocation.

\subsection{Agentic OPD with Prefix Replay}
\label{sec:problem-setup}

\paragraph{OPD in agentic tasks.} We consider tasks $x\sim p(x)$ in a long-horizon environment. At turn $t$, a policy $\pi$ conditions on history $h_t=(x,o_1,a_1,\ldots,o_{t-1},a_{t-1},o_t)$ and samples a complete response $a_t\sim\pi(\cdot\mid h_t)$. The environment returns an observation $o_{t+1}$, which may be terminal. A rollout produces trajectory $\tau=(x,o_1,a_1,\ldots,o_T,a_T,o_{T+1})$.

Let $N_\tau$ be the number of student-generated tokens in $\tau$, and let $s_k$ denote the context before the $k$-th token. Let $\pi_\theta$ and $\pi_T$ denote the current student and fixed teacher, respectively. Fully online agent OPD executes the current student to sample $\tau\sim p_\theta(\cdot\mid x)$, then minimizes the token-averaged reverse KL along the trajectory:
\begin{equation}
    \mathcal{L}_{\mathrm{OPD}}(\theta)
    =\mathbb{E}_{\substack{x\sim p(x),\\ \tau\sim p_\theta(\cdot\mid x)}}
    \left[\frac{1}{N_\tau}\sum_{k=1}^{N_\tau}
    D_{\mathrm{KL}}\!\left(
    \pi_\theta(\cdot\mid s_k)
    \,\middle\|\,
    \pi_T(\cdot\mid s_k)
    \right)\right].
    \label{eq:online-agent-opd}
\end{equation}
Each learner update therefore depends on a fresh rollout in the environment.

\paragraph{Prefix replay.} Prefix replay removes this coupling between environment interaction and policy update. A separate collection stage generates trajectories from fixed policy snapshots and stores their interaction histories. During training, prefix replay truncates a stored trajectory at turn $t$ to obtain $h_{t}$, then samples a response $y\sim\pi_\theta(\cdot\mid h_{t})$. The teacher is evaluated along $y$, but the response is never executed. For a response of length $L$, the replay loss is
\begin{equation}
    \mathcal{L}_{\mathrm{replay}}(\theta;h_{t},y)
    =\frac{1}{L}\sum_{j=1}^{L}
    D_{\mathrm{KL}}\!\left(
    \pi_\theta(\cdot\mid h_{t},y_{<j})
    \,\middle\|\,
    \pi_T(\cdot\mid h_{t},y_{<j})
    \right).
    \label{eq:prefix-replay-loss}
\end{equation}
One replay update generates a response instead of a trajectory, and one executed history can support repeated updates as the student changes. Prefix replay therefore removes environment latency from the learner's critical path. It still leaves open which histories are available and which candidate updates receive training computation.

\subsection{Trajectory Collection and Reuse}
\label{sec:failure-modes}

We first analyze how trajectory collection and reuse determine which replay candidates contribute to student training in prefix replay. Let $\mathcal{D}$ denote the retained trajectory pool, $h_t$ the prefix of $\tau$ before turn $t$, and $\mathcal{B}$ the current replay minibatch. The current student samples $y\sim\pi_\theta(\cdot\mid h_t)$ from this prefix. We decompose the unnormalized training mass of candidate $(\tau,t,y)$ as
\begin{equation}
    \widetilde q_{\theta}(\tau,t,y;\mathcal{B})
    =\underbrace{p_{\mathcal{D}}(\tau)}_{\text{trajectory sampling}}
    \quad
    \underbrace{q_{\mathrm{alloc}}(t\mid\tau)}_{\text{turn allocation}}
    \quad
    \underbrace{\pi_{\theta}(y\mid h_{t})}_{\text{response sampling}}
    \quad
    \underbrace{g(h_{t},y;\mathcal{B})}_{\text{response acceptance}}.
    \label{eq:distillation-turn-mass}
\end{equation}
This formulation exposes two limitations of static prefix policies in long-horizon tasks. First, teacher-only collection for $\mathcal{D}$ misses student-induced histories. Meanwhile, $p_{\mathcal{D}}$ determines the available histories but cannot ensure reliable teacher targets for the responses $y$, and broader history coverage can place the teacher in unfamiliar states. History coverage and response acceptance are therefore designed jointly. Second, $q_{\mathrm{alloc}}$ distributes replay across turns of different importance, where a fixed position rule cannot well identify critical turns to learn.

We address the first limitation by expanding teacher rollouts with student roll-ins, and gate the responses with teacher confidence. We address the second with an allocation rule conditioned on the completed trajectory. CARE-OPD uses the whole trajectory and execution evidence to decide where replay occurs. We detail the two components below.

\subsection{Reliability-Controlled History Coverage}
\label{sec:reliability-controlled-coverage}

Prefix replay is on-policy for the sampled response but off-policy for its stored history. A teacher-only pool omits states caused by earlier student actions, and one early deviation can change many later observations. We extend the candidate history distribution with student-seeded teacher completions. A frozen student snapshot generates an initial trajectory segment, after which the teacher continues from the resulting history.

For each trajectory $i$, we draw $z_i\sim\mathrm{Bernoulli}(\alpha)$, where $\alpha$ is the student-seeded trajectory fraction. If $z_i=0$, the teacher generates the full trajectory. If $z_i=1$, we sample the student roll-in length uniformly as $K_i\sim\mathrm{Uniform}\{1,\ldots,K_{\max}\}$.
The student acts through turn $K_i$, and the teacher completes the trajectory from turn $K_i+1$. Let $\pi_{\theta_0}$ denote the frozen student used for collection. The behavior policy is
\begin{equation}
    \pi_i^{\mathrm{col}}(\cdot\mid h_t)=
    \begin{cases}
        \pi_{\theta_0}(\cdot\mid h_t), & z_i=1\ \text{and}\ t\leq K_i,\\
        \pi_T(\cdot\mid h_t), & \text{otherwise}.
    \end{cases}
    \label{eq:collection-policy}
\end{equation}
The roll-in rate changes the source of collected histories. Student turns remain as context and never become replay targets, while only turns from the teacher continuation are eligible for allocation. If the student completes the task during the roll-in, we discard the trajectory because it contains no teacher continuation for replay.

Expanded history coverage does not guarantee a reliable update. We observe more low-confidence evaluations on the student responses near the switch points of student-seeded trajectories, as analyzed in Section~\ref{sec:coverage-acceptance-analysis}. We therefore decide acceptance after sampling the response. For a candidate prefix $h$ and response $y\sim\pi_\theta(\cdot\mid h)$, let $p^T_j=\pi_T(\cdot\mid h,y_{<j})$ at token position $j$. Inspired by the turn-level confidence weighting in SAGE-OPD~\citep{zhou2026sageopd}, we summarize teacher certainty using the mean top-token probability:
\begin{equation}
    P_T(h,y)=\frac{1}{L}\sum_{j=1}^{L}\max_{v\in\mathcal{V}}p^T_j(v),
    \label{eq:teacher-confidence}
\end{equation}
where $L$ is the response length and $\mathcal{V}$ is the token vocabulary.
For rejection fraction $\gamma\in[0,1]$, let $\tau_{\gamma}(\mathcal{B})$ be the $\gamma$-quantile of confidence scores in minibatch $\mathcal{B}$. We accept the update when
\begin{equation}
    g_{\gamma}(h,y;\mathcal{B})=\mathbb{I}\!\left[P_T(h,y)\geq\tau_{\gamma}(\mathcal{B})\right].
    \label{eq:confidence-gate}
\end{equation}

The gate rejects the lowest-confidence fraction $\gamma$ and keeps the effective batch size predictable. It treats confidence as a proxy for teacher certainty rather than response correctness. A rejected response contributes no gradient, but its prefix remains available for later replay. Because OPD already computes the teacher logits in \Eqref{eq:teacher-confidence}, the gate adds little inference cost.

\subsection{Hindsight Replay Allocation}
\label{sec:hindsight-selection}

Reliability-controlled coverage determines where replay can occur, but long trajectories still contain many candidate turns with limited training value. We allocate replay after each trajectory terminates, when later observations and the final outcome are available. A model-based judge receives the task, complete trajectory, execution feedback, outcome, and a fixed selection budget. It selects teacher-generated turns whose diagnosis, correction, or verification is supported by later evidence.

For trajectory $i$, let $\mathcal{E}_i$ denote the eligible teacher-generated turn indices. Given selection fraction $r$, the judge returns a subset $\mathcal{S}_i\subseteq\mathcal{E}_i$ of size $m_i=\lceil r|\mathcal{E}_i|\rceil$. This defines
\begin{equation}
    q_{\mathrm{alloc}}(t\mid\tau_i)
    =\frac{\mathbb{I}[t\in\mathcal{S}_i]}{m_i}.
    \label{eq:hindsight-mask}
\end{equation}

The fixed within-trajectory fraction avoids score calibration across tasks. During training, we first sample a trajectory and then one of its selected turns, preventing long trajectories from receiving extra weight solely because they contain more turns.

We assign each turn a temporary identifier and append one selection query after the rollout terminates. The query reuses the rollout teacher's trajectory KV cache. We remove the temporary identifiers and selector output before replay, so the student receives the original prefix without future evidence.

\subsection{Training Procedure and Cost}
\label{sec:end-to-end-method}

Algorithm~\ref{alg:care-opd} summarizes the complete CARE-OPD procedure.
Environment interactions occur only during pool construction, and hindsight replay allocation runs once per collected trajectory.
The selected prefixes can support repeated updates as the student changes. Training therefore retains response-level on-policy generation without placing long environment rollouts on the learner's critical path.

\begin{algorithm}[!htbp]
    \caption{CARE-OPD: collection, hindsight allocation, and OPD training.}
    \label{alg:care-opd}
    \small
    \begin{algorithmic}[1]
        \Require Collection task set $\mathcal{X}_{\mathrm{col}}$; $\pi_T$, $\pi_{\theta_0}$, $\alpha$, $K_{\max}$, $r$, $\gamma$; optimization steps $U$, batch size $B$, learning rate $\eta$
        \Ensure Trained student $\pi_\theta$
        \Statex \textbf{Stage I: Collection and hindsight allocation}
        \State $\mathcal{D}\gets\varnothing$
        \For{$x\in\mathcal{X}_{\mathrm{col}}$}
            \State Sample $z\sim\mathrm{Bernoulli}(\alpha)$
            \State If $z=1$, sample $K\sim\mathrm{Uniform}\{1,\ldots,K_{\max}\}$ \Comment{student roll-in length}
            \State Execute $\pi^{\mathrm{col}}$ from \Eqref{eq:collection-policy} on $x$ to obtain $\tau$; let $\mathcal{E}$ be its teacher-generated turns
            \State Discard $\tau$ and \textbf{continue} if verification fails or $\mathcal{E}=\varnothing$ \Comment{retain verified completions}
            \State $\mathcal{S}\gets\Call{HindsightSelect}{\pi_T,x,\tau,\mathcal{E},\lceil r|\mathcal{E}|\rceil}$ \Comment{allocate turns using downstream evidence}
            \State $\mathcal{D}\gets\mathcal{D}\cup\{(\tau,\mathcal{S})\}$
        \EndFor
        \Statex \textbf{Stage II: Replay training}
        \State $\theta\gets\theta_0$
        \For{$u=1,\ldots,U$}
            \State Sample minibatch $\{(\tau_b,\mathcal{S}_b)\}_{b=1}^{B}$ uniformly from $\mathcal{D}$ \Comment{$\mathcal{S}_b$: hindsight-selected turns of $\tau_b$}
            \State Sample $t_b\sim\mathrm{Uniform}(\mathcal{S}_b)$ and extract $h_b$ at $t_b$ for all $b$ \Comment{sample one prefix per trajectory}
            \State Sample $y_b\sim\pi_\theta(\cdot\mid h_b)$ for all $b$
            \State Set $\ell_b\gets\mathcal{L}_{\mathrm{replay}}(\theta;h_b,y_b)$ and $c_b\gets P_T(h_b,y_b)$ for all $b$ \Comment{calculate loss and teacher confidence}
            \State $q_\gamma\gets\Call{Quantile}{\{c_b\}_{b=1}^{B},\gamma}$; $\mathcal{A}\gets\{b:c_b\geq q_\gamma\}$ \Comment{reject bottom-$\gamma$ confidence}
            \State $\displaystyle \theta\gets\theta-\eta\nabla_\theta\frac{1}{|\mathcal{A}|}\sum_{b\in\mathcal{A}}\ell_b$
        \EndFor
        \State \Return $\pi_\theta$
    \end{algorithmic}
\end{algorithm}

\FloatBarrier

\section{Experiments}
\label{sec:experiments}

\subsection{Experimental Setup}
\label{sec:experimental-setup}

\paragraph{Data and benchmarks.} We study two long-horizon agentic tasks: Software Engineering (SWE) and Deep Research. For \textbf{Software Engineering}, we partition SWE-Gym~\citep{pan2025swegym} into a training set and a validation set of 100 tasks. We collect all training trajectories from the training set and evaluate task success on SWE-bench Verified~\citep{jimenez2024swebench}. For \textbf{Deep Research}, as high-quality data is important, we use a dataset curated from public sources with a quality-control pipeline, detailed in Appendix~\ref{app:deep-research-data}. We evaluate answer accuracy on BrowseComp-Plus~\citep{chen2026browsecompplus}, using Qwen3-Embedding-8B as the retriever and gpt-oss-20b as the judge.

\paragraph{Agent scaffolds.} For SWE, we use OpenHands as the agent scaffold and retain its default system prompt and tools~\citep{wang2025openhands}. For Deep Research, we customize the Uni-Agent scaffold~\citep{ding2026uniagent} and provide three tools: \texttt{search}, \texttt{open}, and \texttt{find}. SFT and OPD training both use a 128K context limit and a 150-turn limit.

\paragraph{Implementation details.} For OPD training, we use 4 GPU nodes, where 1 node is for the teacher, 2 nodes are for student rollouts, and 1 node is for student updates. We optimize for 60 steps. Appendix~\ref{app:training-hyperparameters} lists the complete training configuration. For each training method, we train the policy with 3 seeds and report mean\,$\pm$\,sample SD.

\begin{table}[!htbp]
    \centering
    \small
    \setlength{\tabcolsep}{8pt}
    \caption{Task success (\%) on SWE-bench Verified and answer accuracy (\%) on BrowseComp-Plus. All training methods start from the same initial student. We omit Qwen3 Deep Research experiments because its candidate teacher exceeds the initial student by only 4.6 points on BrowseComp-Plus, leaving limited room for distillation; Appendix~\ref{app:qwen3-deep-research} reports this assessment.}
    \label{tab:main-results}
    \begin{tabular}{@{}lccc@{}}
        \toprule
        & \multicolumn{2}{c}{\emph{Qwen3.5 family}} & \multicolumn{1}{c}{\emph{Qwen3 family}} \\
        \cmidrule(lr){2-3}\cmidrule(lr){4-4}
        Method & \shortstack{SWE-bench\\Verified$\uparrow$} & \shortstack{BrowseComp-\\Plus$\uparrow$} & \shortstack{SWE-bench\\Verified$\uparrow$} \\
        \midrule
        \rowcolor{familyrow}[0pt][0pt]
        \multicolumn{4}{@{}l@{}}{\emph{\strut Reference models}} \\
        \addlinespace[1pt]
        Teacher & 67.2 & 57.8 & 43.0 \\
        Initial student & 53.2 & 42.5 & 10.8 \\
        \addlinespace[1pt]
        \rowcolor{familyrow}[0pt][0pt]
        \multicolumn{4}{@{}l@{}}{\emph{\strut Offline teacher imitation}} \\
        \addlinespace[1pt]
        SFT & \mbox{54.7\smallsd{0.4}} & \mbox{46.1\smallsd{1.5}} & \mbox{17.6\smallsd{1.0}} \\
        \addlinespace[1pt]
        \rowcolor{familyrow}[0pt][0pt]
        \multicolumn{4}{@{}l@{}}{\emph{\strut Online OPD}} \\
        \addlinespace[1pt]
        Full trajectory (all turns) & \mbox{56.3\smallsd{0.6}} & \mbox{\underline{47.2}\smallsd{0.6}} & \mbox{18.7\smallsd{0.8}} \\
        One random turn & \mbox{\underline{57.7}\smallsd{0.3}} & \mbox{47.1\smallsd{0.2}} & \mbox{\underline{20.3}\smallsd{0.5}} \\
        \addlinespace[1pt]
        \rowcolor{familyrow}[0pt][0pt]
        \multicolumn{4}{@{}l@{}}{\emph{\strut Prefix replay}} \\
        \addlinespace[1pt]
        Vanilla prefix replay & \mbox{53.7\smallsd{1.1}} & \mbox{45.9\smallsd{1.3}} & \mbox{17.1\smallsd{0.8}} \\
        ReOPD (position decay) & \mbox{53.9\smallsd{0.2}} & \mbox{45.5\smallsd{0.9}} & \mbox{18.3\smallsd{1.0}} \\
        \textbf{CARE-OPD (ours)} & \mbox{\textbf{59.5}\smallsd{0.3}} & \mbox{\textbf{49.6}\smallsd{0.5}} & \mbox{\textbf{23.0}\smallsd{0.5}} \\
        \bottomrule
    \end{tabular}
\end{table}

\paragraph{Models.} We conduct experiments using two model families, Qwen3.5 and Qwen3. For \textbf{Qwen3.5}, Qwen3.5-9B is the student and Qwen3.5-27B is the teacher, with thinking enabled. For \textbf{Qwen3}, Qwen3-8B is the student and Qwen3-Coder-30B-A3B-Instruct is the teacher. Qwen3-8B and Qwen3-Coder-30B-A3B-Instruct use different native tool-call formats, making direct OPD ineffective, so we first cold-start Qwen3-8B as detailed in Appendix~\ref{app:qwen3-cold-start}. We report the cold-start model as the Qwen3 initial student and use it to initialize every Qwen3 training method. For Qwen3 on BrowseComp-Plus, the teacher has a small performance gap relative to the initial student and Appendix~\ref{app:qwen3-deep-research} reports the scores, so we omit experiments for this setting.

\paragraph{Baselines.} All methods within each model family start from the same initial checkpoint. The \textbf{SFT} baseline behavior-clones successful teacher trajectories retained by rejection sampling. For Deep Research, we generate candidate SFT trajectories from a fixed sample of 3,840 tasks, matching the task budget of each OPD run. \textbf{Fully online OPD} executes the current student in the environment and applies the OPD loss. \textbf{ReOPD} uses teacher-generated histories and position-decayed replay with the best-performing value in our sweep, $\kappa=0.9$, and Appendix~\ref{app:construction-ablations} reports the full sweep. \textbf{Vanilla prefix replay} uses teacher-generated histories and samples eligible turns uniformly, without hindsight replay allocation or confidence gating.

\subsection{Main Results}
\label{sec:main-results}

\paragraph{CARE-OPD outperforms both online OPD and existing prefix replay methods.} This result holds across all three settings, showing that adaptive history construction and replay allocation allow environment-decoupled replay to surpass training on fresh online trajectories. The following paragraphs analyze these comparisons in detail.

\paragraph{One-turn online OPD is comparable to full-trajectory OPD, but fresh student histories matter.} The two online variants differ by only 1.4, 0.1, and 1.6 points across the three settings, with no consistent winner. On Qwen3.5 SWE-bench Verified, trajectories average 84.9 turns, yet random-turn OPD scores 1.4 points higher. Less reliable targets after accumulated student errors in long trajectories may limit the value of full-trajectory distillation. However, vanilla prefix replay trails one-turn online OPD by 4.0, 1.2, and 3.2 points. This contrast points to fixed teacher-generated prefixes and absent environment execution, rather than one-turn supervision itself: replay cannot reach later histories induced by the current student's decisions. Section~\ref{sec:wall-clock-efficiency} below shows the trade-off: vanilla prefix replay is substantially cheaper but less effective.

\paragraph{Position-based replay allocation yields limited gains in long trajectories.} Relative to vanilla prefix replay, ReOPD changes performance by +0.2, -0.4, and +1.2 points. Its monotone schedule assumes that useful supervision concentrates near the trajectory start. Long-horizon agents also make consequential diagnoses, corrections, and verification decisions at later turns, so fixed decay can suppress useful examples. Section~\ref{sec:allocation-effectiveness} examines this limitation directly. Together, the online and replay comparisons show that more supervision is not consistently better: full-trajectory OPD does not reliably beat one-turn OPD, and position decay does not reliably beat uniform replay. CARE-OPD instead adapts history coverage and selects turns from trajectory evidence, consistent with its gains over both replay variants.

\FloatBarrier

\subsection{Does Prefix Replay Improve End-to-End Training Efficiency?}
\label{sec:wall-clock-efficiency}

We compare synchronous online OPD, asynchronous online OPD, vanilla prefix replay, and CARE-OPD under identical GPU, environment worker, and batch size configurations. Appendix~\ref{app:training-hyperparameters} reports the detailed hyperparameters. End-to-end time includes every stage needed to obtain the trained student. For CARE-OPD, this consists of trajectory collection, hindsight replay allocation, and replay training. Environment actions count all attempted trajectories, including unsuccessful collection attempts.

\begin{table}[!htbp]
    \centering
    \small
    \setlength{\tabcolsep}{8pt}
    \caption{SWE end-to-end training efficiency and task success. Task success is reported in percent. Parentheses give ratios relative to synchronous online OPD with one distilled turn.}
    \label{tab:wall-clock}
    \setlength{\tabcolsep}{6pt}
    \begin{tabular}{@{}lccc@{}}
        \toprule
        Method & E2E time (h)$\downarrow$ & Env. actions (k)$\downarrow$ & Task success$\uparrow$ \\
        \midrule
        Sync OPD, one turn & \mbox{7.5\smallsd{0.5}}\ratio{1.0} & \mbox{309.9\smallsd{5.4}}\ratio{1.0} & \mbox{\underline{57.7}\smallsd{0.3}} \\
        Async OPD, one turn & \mbox{5.5\smallsd{0.1}}\ratio{0.7} & \mbox{308.1\smallsd{3.4}}\ratio{1.0} & \mbox{57.1\smallsd{0.4}} \\
        Async OPD, all turns & \mbox{5.8\smallsd{0.1}}\ratio{0.8} & \mbox{314.8\smallsd{8.9}}\ratio{1.0} & \mbox{56.5\smallsd{1.4}} \\
        Vanilla prefix replay & \mbox{\textbf{2.9}\smallsd{0.2}}\ratiob{0.4} & \mbox{\textbf{169.5}\smallsd{0.9}}\ratiob{0.5} & \mbox{53.7\smallsd{1.1}} \\
        CARE-OPD & \mbox{\underline{3.0}\smallsd{0.1}}\ratio{0.4} & \mbox{\underline{171.2}\smallsd{3.3}}\ratio{0.6} & \mbox{\textbf{59.5}\smallsd{0.3}} \\
        \bottomrule
    \end{tabular}
\end{table}

Table~\ref{tab:wall-clock} shows that asynchronous one-turn OPD reduces end-to-end time by 27\% relative to synchronous one-turn OPD while achieving comparable task success (57.1\% vs.\ 57.7\%). However, it executes nearly the same number of environment actions (308.1k vs.\ 309.9k) and retains the latency tail of full environment rollouts. By decoupling environment interaction from student training, vanilla prefix replay and CARE-OPD reduce end-to-end time by 61\% and 60\% relative to synchronous one-turn OPD, and by 47\% and 45\% relative to asynchronous one-turn OPD. Both replay methods also execute approximately 45\% fewer environment actions than either online baseline because each collected trajectory supports repeated prefix updates. CARE-OPD takes 3.0 hours and executes 171.2k environment actions, only 0.1 hours and 1.7k actions more than vanilla prefix replay, while improving task success by 5.8 points. Its evidence-adaptive components therefore add little cost to vanilla prefix replay.

\FloatBarrier

\subsection{How Do History Coverage and Response Acceptance Interact?}
\label{sec:coverage-acceptance-analysis}

\begin{figure}[!htbp]

    \centering
    \includegraphics[width=0.45\linewidth]{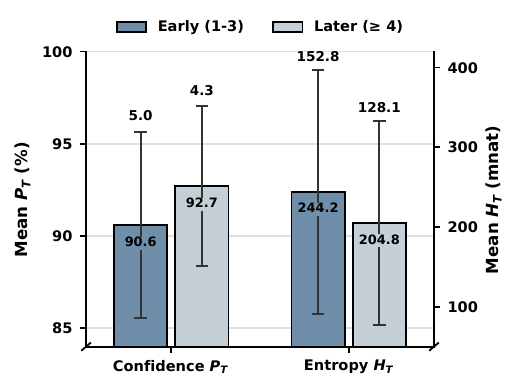}
    \setlength{\belowcaptionskip}{7pt}

    \caption{Teacher uncertainty after student-to-teacher switch points.}
    \label{fig:post-roll-in-uncertainty}

\end{figure}

Student-seeded collection broadens history coverage, but sampled responses on the resulting prefixes can expose the teacher to states outside its usual interaction distribution. We first measure teacher uncertainty around the roll-in switch, then compare training outcomes across history pools and gate choices.

\paragraph{Measure teacher uncertainty.} We examine teacher uncertainty immediately after the student-to-teacher switch on collected trajectories with student roll-in. For each response, we report the mean top-token probability $P_T$ from \Eqref{eq:teacher-confidence} and the token-averaged teacher entropy $H_T=-\frac{1}{L}\sum_{j=1}^{L}\sum_{v\in\mathcal{V}}p^T_j(v)\log p^T_j(v)$.

Figure~\ref{fig:post-roll-in-uncertainty} summarizes the shift in the first three turns after the switch. Relative to later turns, the first three turns have a lower mean $P_T$ by 2.1 points and raise the mean $H_T$ by 39.4 mnat. The early low-confidence ($<90\%$) and high-entropy ($>400$ mnat) rates are 35.1\% and 16.5\%, compared with 26.2\% and 9.3\% over all post-switch turns. This localized shift motivates the response-level reliability control.

\begin{table}[!htbp]

    \centering
    \small
    \setlength{\belowcaptionskip}{4pt}
    \caption{Interaction between history coverage and response acceptance on SWE-bench Verified. We report task success (\%).}
    \label{tab:coverage-acceptance-interaction}
    \setlength{\tabcolsep}{10pt}
    \renewcommand{\arraystretch}{1.1}
    \begin{tabular}{@{}lcc@{}}
        \toprule
        History pool & \shortstack{No confidence\\gate} & \shortstack{Confidence\\gate} \\
        \midrule
        Teacher-only & \mbox{53.7\smallsd{1.1}} & \mbox{54.5\smallsd{0.8}} \\
        Student-seeded & \mbox{55.1\smallsd{0.5}} & \mbox{\textbf{57.1}\smallsd{1.5}} \\
        \bottomrule
    \end{tabular}

\end{table}

\paragraph{Confidence gating complements expanded history coverage.} We next measure how history coverage and response acceptance affect downstream training. Table~\ref{tab:coverage-acceptance-interaction} forms a $2\times2$ comparison between teacher-only and student-seeded trajectory pools, with or without confidence gating. The student-seeded pool uses our default fraction $\alpha=0.1$.
Student roll-in with gating reaches 57.1 points, a 3.4-point gain over the baseline. This gain exceeds the 2.2-point sum of the two standalone gains by 1.2 points, indicating a positive interaction. Confidence gating is therefore more useful on the expanded history support, consistent with the elevated teacher uncertainty after student roll-in.

\subsection{Hindsight Replay Allocation Improves Prefix Replay}
\label{sec:allocation-effectiveness}

\begin{table}[!htbp]

    \centering
    \small
    \setlength{\belowcaptionskip}{4pt}
    \caption{Controlled comparison of replay-allocation rules on SWE-bench Verified.}
    \label{tab:level-controls}
    \setlength{\tabcolsep}{4pt}
    \renewcommand{\arraystretch}{1.1}
    \begin{tabular}{@{}lcc@{}}
        \toprule
        Allocation rule & \shortstack{Retained\\turns} & \shortstack{Task success\\(\%)} \\
        \midrule
        Uniform replay & 100\% & \mbox{53.7\smallsd{1.1}} \\
        Random subset & 40\% & \mbox{53.7\smallsd{0.4}} \\
        Best position decay (ReOPD) & weighted & \mbox{53.9\smallsd{0.2}} \\
        Separate critic & 40\% & \mbox{\textbf{57.6}\smallsd{2.0}} \\
        Teacher hindsight & 40\% & \mbox{57.3\smallsd{0.8}} \\
        \bottomrule
    \end{tabular}

\end{table}

We vary replay allocation strategies. Table~\ref{tab:level-controls} compares five allocation rules. Uniform replay samples all eligible turns, while random subset selection retains 40\% of the turns in each trajectory without using their content. Best position decay uses the best-performing ReOPD setting in our sweep ($\kappa=0.9$) in Appendix~\ref{app:construction-ablations}. The separate critic uses a stronger critic model Qwen3.5-122B-A10B for hindsight replay allocation, whereas our judge appends one selection query to the rollout teacher and reuses its trajectory KV cache. Both model-based selectors use the same 40\% per-trajectory budget.

\paragraph{Trajectory evidence drives the allocation gain.} The hindsight allocation in CARE-OPD improves task success by 3.6 points over uniform replay and by 3.6 points over the matched random subset. Random subset selection is comparable with uniform replay, so retaining fewer turns alone does not improve performance. Best position decay improves over uniform replay by 0.2 points but remains 3.4 points below our allocation. These results indicate that selecting turns using trajectory evidence accounts for the gain, while a monotone preference for early turns yields only a marginal improvement on these long trajectories.

\begin{figure}[!htbp]

    \centering
    \includegraphics[width=0.45\linewidth]{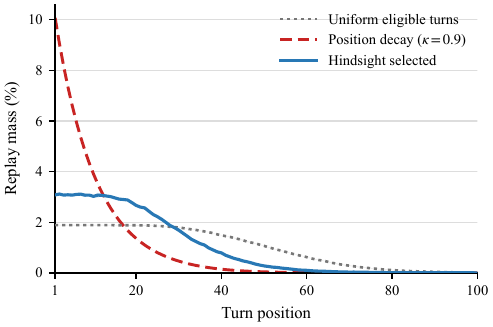}

    \caption{Replay mass by turn position over the first 100 turns. }
    \label{fig:turn-position}

\end{figure}

\paragraph{Hindsight allocation preserves useful late turns.} Figure~\ref{fig:turn-position} visualizes the normalized replay mass. The step-decay rule assigns most replay mass to the first few turns, whereas hindsight replay allocation spreads replay across the first several dozen turns and retains a longer tail. These later turns include corrections, recovery actions, and verification steps supported by subsequent execution evidence. Their value depends on what happened in the trajectory rather than on a fixed depth schedule.

\paragraph{The rollout teacher nearly matches a separate critic.} The separate critic reaches 57.6\%, only 0.3 points above our 57.3\%. Reusing the rollout teacher therefore nearly matches the stronger independent critic while avoiding an additional full-trajectory prefill. Appendices~\ref{app:hyperparameter-sensitivity} and~\ref{app:construction-ablations} report the hyperparameter and position-decay sweeps, respectively.

\FloatBarrier

\section{Limitations and Conclusion}

\paragraph{Limitations.} Although our student-seeded collection broadens history coverage, it uses a frozen initial student and bounded roll-ins, while the student changes during training. The replay pool may therefore become less representative over time. Periodically refreshing the pool could reduce this mismatch at additional environment cost. Replayed responses are not executed, leaving their validity and environmental consequences unobserved. Teacher confidence measures distributional certainty rather than response correctness. Future work could selectively execute uncertain responses or develop execution-aware correctness estimators.

\paragraph{Conclusion.} On-policy distillation is expensive for long-horizon agents because each update typically requires a fresh environment rollout. We study prefix replay as an efficient alternative, and our analysis identifies two central design problems: trajectory collection and trajectory reuse. We introduce CARE-OPD, which addresses trajectory collection with reliability-controlled history coverage and trajectory reuse with evidence-based allocation. CARE-OPD outperforms online OPD and prior prefix replay methods across long-horizon tasks SWE and Deep Research. On SWE-bench Verified, it achieves 59.5\% task success with 45\% less end-to-end training time and 44\% fewer environment actions than asynchronous online OPD. Further analyses show that adaptive collection and reuse make prefix replay an effective and efficient training approach for long-horizon agents.

\clearpage

\FloatBarrier
\bibliographystyle{ieeenat_fullname}
\bibliography{main}

\clearpage
\appendix

\section*{Appendix}

\section{Deep Research Dataset Construction}
\label{app:deep-research-data}

We construct the Deep Research training-task pool from five public sources. We combine OpenResearcher~\citep{li2026openresearcher}, DeepForge from OffSeeker~\citep{zhou2026offseeker}, OpenSeeker~\citep{du2026openseeker}, REDSearcher~\citep{chu2026redsearcher} and public Deep Research SFT collections released by Kyle Montgomery~\citep{montgomery2026deepresearch}. Source releases may contain complete agent trajectories, but we retain only the query and its reference answer from each example. We discard every source trajectory and never use it as a training target.

We normalize each query with Unicode NFKC normalization, case folding, and whitespace collapsing. We then merge the sources and deduplicate queries across all sources. This process yields 75,372 queries.

Because the source corpora were collected with different pipelines and at different times, their query--answer pairs vary in quality and current verifiability. We use DeepSeek-V4-Flash~\citep{deepseekai2026deepseekv4} as a task-level answerability probe in a common live-web environment. The probe can call \texttt{search}, \texttt{open}, and \texttt{find}, while web search and reading use Serper\footnote{\url{https://serper.dev/}}. We retain a candidate when the probe's final answer is consistent with the source reference and supported by evidence available in the current web environment. This pass screens for ambiguous queries, stale reference answers, and questions whose supporting evidence is no longer accessible. It also reduces variation caused by source-specific retrieval environments. After filtering, the remaining 16,855 tasks form the curated training dataset for Deep Research.

\section{Training Hyperparameters}
\label{app:training-hyperparameters}

All OPD-based methods use four GPU nodes, each equipped with eight NVIDIA H200 GPUs. We assign one node to teacher inference, two nodes to student rollout generation, and one node to student updates. Across these methods, we use a learning rate of $10^{-6}$, a batch size of 64, and 60 optimization steps.

For asynchronous OPD, we set the maximum policy staleness to 4 learner updates. We observe that staleness values at most 4 have little effect on final task performance, whereas larger values reduced performance. For SFT, we use a learning rate of $10^{-5}$, a batch size of 64, and train for three epochs. Table~\ref{tab:training-hyperparameters} summarizes these settings.

\begin{table}[!htbp]
    \centering
    \small
    \setlength{\tabcolsep}{10pt}
    \caption{Training hardware and optimization hyperparameters.}
    \label{tab:training-hyperparameters}
    \begin{tabular}{@{}lcc@{}}
        \toprule
        Configuration & OPD & SFT \\
        \midrule
        Learning rate & $10^{-6}$ & $10^{-5}$ \\
        Batch size & 64 & 64 \\
        Training duration & 60 steps & 3 epochs \\
        Maximum staleness & 4 (asynchronous OPD only) & -- \\
        \bottomrule
    \end{tabular}
\end{table}

\section{Qwen3 Cold Start}
\label{app:qwen3-cold-start}

Qwen3-8B and Qwen3-Coder-30B-A3B-Instruct use different native tool-call formats. The raw student therefore produces invalid calls under the Coder format, making direct OPD ineffective. We randomly sample 500 tasks from R2E-Gym-Subset~\citep{jain2025r2egym}, whose repositories do not overlap with SWE-bench, and use Qwen3-Coder-30B-A3B-Instruct to generate candidate trajectories. We retain successful trajectories through rejection sampling and perform SFT on Qwen3-8B using the retained data. The raw Qwen3-8B model scores 8.6\% on SWE-bench Verified, while the cold-start model scores 10.8\%. We report the cold-start model as the Qwen3 initial student and use it to initialize every Qwen3 training method.

\section{Qwen3 Deep-Research Assessment}
\label{app:qwen3-deep-research}

We evaluated the Qwen3 teacher and the same cold-start initial student on BrowseComp-Plus. Table~\ref{tab:qwen3-deep-research} shows that Qwen3-Coder-30B-A3B-Instruct reaches 17.3\% accuracy, while Qwen3-8B reaches 12.7\%. The teacher is only 4.6 percentage points higher than the student. Thus, we exclude this setting from our experiments.

\begin{table}[!htbp]
    \centering
    \small
    \setlength{\tabcolsep}{8pt}
    \caption{Answer accuracy on BrowseComp-Plus for the candidate Qwen3 teacher and initial student.}
    \label{tab:qwen3-deep-research}
    \begin{tabular}{@{}llc@{}}
        \toprule
        Model & Role & Accuracy (\%) \\
        \midrule
        Qwen3-Coder-30B-A3B-Instruct & Candidate teacher & 17.3 \\
        Qwen3-8B & Initial student & 12.7 \\
        \bottomrule
    \end{tabular}
\end{table}

\section{Sensitivity to the ReOPD Position-Decay Parameter}
\label{app:construction-ablations}

We sweep the position-decay parameter $\kappa$ on SWE-bench Verified using the same collection and attempted-update budgets across settings. For the eligible turn set $\mathcal{E}_\tau$ of trajectory $\tau$, we apply the ReOPD schedule to the absolute turn index and sample turns according to
\begin{equation}
    q_{\mathrm{alloc}}(t\mid\tau)
    =\frac{\kappa^t}{\sum_{j\in\mathcal{E}_\tau}\kappa^j},
    \qquad t\in\mathcal{E}_\tau.
    \label{eq:reopd-position-decay}
\end{equation}
At $\kappa=1.0$, this schedule assigns equal probability to all eligible turns (vanilla prefix replay).
\begin{table}[!htbp]
    \centering
    \small
    \setlength{\tabcolsep}{10pt}
    \caption{Sensitivity to the position-decay parameter $\kappa$ on SWE-bench Verified. The setting $\kappa=1$ is uniform replay.}
    \label{tab:allocation-sensitivity}
    \begin{tabular}{@{}lcccc@{}}
        \toprule
        $\kappa$ & 0.5 & 0.7 & 0.9 & 1.0 \\
        Success & \mbox{52.9\smallsd{1.3}} & \mbox{53.9\smallsd{0.4}} & \mbox{\textbf{53.9}\smallsd{0.2}} & \mbox{53.7\smallsd{1.1}} \\
        \bottomrule
    \end{tabular}
\end{table}

For $\kappa\in\{0.7,0.9,1.0\}$, the configuration means in Table~\ref{tab:allocation-sensitivity} span 0.2 points. Strong decay degrades performance: reducing $\kappa$ to 0.5 lowers task success to 52.9\%, 1.0 points below the best position-decay setting. At $\kappa=0.5$, the replay weight halves at every turn, so a prefix ten turns later receives only $0.5^{10}\approx0.1\%$ of the weight assigned to the earlier prefix. This rapid decay concentrates replay on the start of a long trajectory and leaves little training mass for later turns, which likely contributes to the observed decline.

\section{Sensitivity to Key Hyperparameters}
\label{app:hyperparameter-sensitivity}

We evaluate three hyperparameters on SWE-bench Verified: the student-seeded trajectory fraction $\alpha$, rejection fraction $\gamma$, and hindsight selection fraction $r$. Each sweep varies one hyperparameter, and holds the remaining settings fixed. The selected settings are $\alpha=0.1$, $\gamma=0.1$, and $r=0.4$. We do not tune the maximum student roll-in length $K_{\max}$. Instead, we set it to the mean trajectory length observed during collection so that the roll-in range scales with the task horizon. For Qwen3.5, this rule gives $K_{\max}=72$ on SWE and $K_{\max}=20$ on Deep Research.

\begin{table}[!htbp]
    \centering
    \small
    \setlength{\tabcolsep}{10pt}
    \renewcommand{\arraystretch}{1.05}
    \caption{Sensitivity of CARE-OPD on SWE-bench Verified. Each block reports the sweep settings and corresponding task success for one hyperparameter. Bold marks the settings used in the final method.}
    \label{tab:key-hyperparameter-sensitivity}
    \begin{tabular}{@{}lcc@{}}
        \toprule
        Hyperparameter & Sweep setting & Task success (\%) \\
        \midrule
        \multirow{3}{*}{Student-seeded fraction $\alpha$} & \textbf{0.1} & \mbox{\textbf{59.5}\smallsd{0.3}} \\
        & 0.3 & \mbox{59.1\smallsd{0.3}} \\
        & 0.5 & \mbox{58.8\smallsd{1.3}} \\
        \midrule
        \multirow{3}{*}{Rejection fraction $\gamma$} & \textbf{0.1} & \mbox{\textbf{59.5}\smallsd{0.3}} \\
        & 0.3 & \mbox{58.9\smallsd{0.7}} \\
        & 0.5 & \mbox{57.7\smallsd{0.6}} \\
        \midrule
        \multirow{4}{*}{Hindsight fraction $r$} & 0.2 & \mbox{57.5\smallsd{0.5}} \\
        & \textbf{0.4} & \mbox{\textbf{59.5}\smallsd{0.3}} \\
        & 0.6 & \mbox{59.3\smallsd{0.9}} \\
        & 0.8 & \mbox{58.7\smallsd{0.9}} \\
        \bottomrule
    \end{tabular}
\end{table}

\FloatBarrier

\textbf{Performance is stable across the tested student-seeding fractions.} Mean task success is 59.5\%, 59.1\%, and 58.8\% at $\alpha=0.1$, 0.3, and 0.5, respectively. These configuration means span 0.7 points over the tested interval. We use $\alpha=0.1$, which gives the highest task success while requiring student roll-ins for only 10\% of the collected trajectories.

\textbf{Rejecting a small low-confidence fraction is sufficient.} Task success decreases markedly as $\gamma$ increases. Because the gate removes the lowest-confidence fraction of each minibatch, increasing $\gamma$ progressively discards more candidate updates. This decline suggests that most of the benefit comes from filtering a small set of potentially unreliable updates, while stronger rejection removes useful supervision.

\textbf{Hindsight allocation benefits from a moderate selection budget.} Selecting 20\% of the eligible turns yields 57.5\% mean task success, while $r=0.4$ and $r=0.6$ reach 59.5\% and 59.3\%, respectively. Performance decreases slightly to 58.7\% at $r=0.8$. With $r=0.2$, the judge selects few turns and therefore covers a narrower set of trajectory states. We also observe that small selection fractions concentrate selected turns near trajectory starts, which can omit later corrections, recovery actions, and verification steps.

\FloatBarrier
\end{document}

%% file: preamble.tex
\usepackage{amsmath,amssymb}
\usepackage{fix-cm}
\usepackage{array}
\usepackage{colortbl}
\usepackage{algorithm}
\usepackage{algpseudocode}

\definecolor{familyrow}{RGB}{242,242,242}
\newcommand{\ratio}[1]{{\,\scriptsize($#1\times$)}}
\newcommand{\ratiob}[1]{{\,\scriptsize($\mathbf{#1\times}$)}}
\newcommand{\Eqref}[1]{Equation~\ref{#1}}

\makeatletter
\newcommand{\smallsd}[1]{{\fontsize{\dimexpr\f@size pt*2/3\relax}{\f@baselineskip}\selectfont\kern0.3pt\textpm\kern0.1pt#1}}
\makeatother

\crefname{figure}{Fig.}{Figs.}
\Crefname{figure}{Fig.}{Figs.}
\crefname{table}{Table}{Tables}
\Crefname{table}{Table}{Tables}

